\documentclass[12pt]{article}

\usepackage{amsmath,amssymb,amsfonts}
\usepackage{bbm}
\usepackage[justification=centering,margin=10pt,font=small,labelfont=bf]{caption}
\usepackage{graphicx,xcolor}

\usepackage{apacite}
\usepackage[comma,authoryear,round]{natbib}
\begin{document}

\noindent From unbiased MDI Feature Importance to Explainable AI for Trees
\vskip 3mm

\vskip 5mm
\noindent Markus Loecher

\noindent  Berlin School of Economics and Law, 

\noindent 10825 Berlin, Germany

\noindent markus.loecher@hwr-berlin.de

\vskip 3mm

\noindent ABSTRACT

We attempt to give a unifying view of the various recent attempts to (i) improve the interpretability of tree-based models and (ii) debias the the default variable-importance measure in random Forests, Gini importance.
In particular, we demonstrate a common thread among the out-of-bag based bias correction methods and their connection to local explanation for trees. In addition, we point out a bias caused by the inclusion of inbag data in the newly developed SHAP values and suggest a remedy.



\section{Variable importance in trees}

Variable importance is not very well defined as a concept. 
Even for the case of a linear model with $n$ observations, $p$ variables and the standard $n >> p$ situation, there is no theoretically defined variable importance metric in the sense of a parametric quantity that a variable importance
estimator should try to estimate \citep{gromping2009variable}.
Variable importance measures for random forests have been receiving increased attention in bioinformatics, for instance to select a subset of genetic markers relevant for the prediction of a certain disease.
They also have been used as screening tools \citep{diaz2006gene,menze2009comparison}
in important applications highlighting the need for reliable and well-understood feature importance
measures.

The default choice in most software implementations \citep{randomForest2002,pedregosa2011scikit} of random forests \citep{Breiman2001} is the \textit{mean decrease in impurity (MDI)}. The MDI of a feature is computed as a (weighted) mean of the individual trees' improvement
in the splitting criterion produced by each variable. A substantial shortcoming of this default measure is its evaluation on the in-bag samples which can lead to severe overfitting \citep{kim2001classification}. It was also pointed out by \cite{Strobl2007a} 
that \textit{the variable importance measures of Breiman's original Random Forest method ... are not reliable in situations where potential predictor variables vary in their scale of measurement or their number of categories}.\\
There have been multiple attempts at correcting the well understood bias of the Gini impurity measure both as a split criterion as well as a contributor to importance scores, each one coming from a different perspective. \\
\cite{strobl2007unbiased} derive the exact distribution of the maximally selected Gini gain along with their resulting p-values by means of a combinatorial approach. 
\cite{shih2004variable} suggest a solution to the bias  for the case of regression trees as well as binary classification trees \citep{shih2004note} which is also based on p-values. Several authors \citep{loh1997split, hothorn2006unbiased} argue that the criterion for split variable and split point selection should be separated.

A different approach is to add so-called pseudo variables to a dataset, which are permuted versions of the original variables and can be used to correct for bias \citep{sandri2008bias}. 
Recently,  a modified version of the Gini importance called Actual Impurity Reduction (AIR) was proposed \cite{nembrini2018revival} that is faster than the original method proposed by Sandri and Zuccolotto with almost no overhead over the creation of the original RFs and available in the R package \textit{ranger}  \citep{wright2015ranger,wright2017package}.

\section{Separating inbag and out-of-bag (oob) samples \label{sec:OOB}}

An idea that is gaining quite a bit of momentum is to include OOB samples to compute a debiased version of the Gini importance \citep{NIPS2019_9017,zhou2019unbiased,loecher2020unbiased} yielding promising results. 
Here, the original Gini impurity (for node $m$) for a categorical variable $Y$ which can take $D$ values $c_1, c_2, \ldots, c_D$  is defined as 
\[
G(m) =  \sum_{d=1}^D{ \hat{p}_d(m) \cdot \left(1- \hat{p_d}(m) \right) }, \text{ where} \hat{p}_d = \frac{1}{n_m} \sum_{i \in m}{Y_i}.
\]
\cite{loecher2020unbiased} proposed a \textit{penalized Gini impurity} which combines inbag and out-of-bag samples. The main idea is to increase the impurity $G(m)$  by a  penalty that is proportional to the difference $\Delta=(\hat{p}_{OOB} - \hat{p}_{inbag})^2$:
 \begin{equation}
 \label{eq:PGoob}
     PG_{oob}^{\alpha,\lambda} = \alpha \cdot G_{oob}   + (1-\alpha) \cdot  G_{in}  + \lambda \cdot (\hat{p}_{oob} - \hat{p}_{in})^2
 \end{equation}
In addition,  \cite{loecher2020unbiased} investigated replacing $G(m)$ by an unbiased estimator of the variance  via the well known sample size correction.
\begin{equation}
\widehat{G}(m)  =  \frac{N}{N-1} \cdot  G(m)  \label{eq:Gmod}
\end{equation}
In this paper we focus on the following three special cases,  [$\alpha=1, \lambda=2$], [$\alpha=0.5, \lambda=1$] as well as [$\alpha=1, \lambda=0$]: 
\begin{align}
PG_{oob}^{(1,2)} & = \sum_{d=1}^D{ \hat{p}_{d,oob} \cdot \left(1- \hat{p}_{d,oob} \right)    + 2  (\hat{p}_{d,oob} - \hat{p}_{d,in})^2} \label{eq:PG1} \\
PG_{oob}^{(0.5,1)} & =  \frac{1}{2} \cdot \sum_{d=1}^D{ \hat{p}_{d,oob} \cdot \left(1- \hat{p}_{d,oob} \right)    +  \hat{p}_{d,in} \cdot \left(1- \hat{p}_{d,in} \right)    + (\hat{p}_{d,oob} - \hat{p}_{d,in})^2} \label{eq:PG2}\\
\widehat{PG}_{oob}^{(1,0)}  & =  \frac{N}{N-1} \cdot  \sum_{d=1}^D{ \hat{p}_{d,oob} \cdot \left(1- \hat{p}_{d,oob} \right) } \label{eq:PG0mod} \\
\end{align}
Our main contributions are to show that 
\begin{itemize}
  \item $PG_{oob}^{(1,2)}$ is equivalent to the \textit{MDI-oob} measure defined in \cite{NIPS2019_9017}.
  \item $PG_{oob}^{(1,2)}$ has close connections to the \textit{conditional feature contributions} (CFCs) defined in  \citep{saabas2019treeInterpreter},  
  \item Similarly to MDI, both the CFCs as well as the related  \textit{SHapley Additive exPlanation} (SHAP) values defined in \citep{lundberg2020local} are susceptible to ``overfitting'' to the training data.
  \item $PG_{oob}^{(0.5,1)}$ is equivalent to the \textit{unbiased split-improvement} measure defined in \cite{zhou2019unbiased}.
\end{itemize}
We refer the reader to \citep{loecher2020unbiased} for a proof that $\widehat{PG}_{oob}^{(1,0)}$ and $PG_{oob}^{(0.5,1)}$ are unbiased estimators of feature importance in the case of non-informative variables.

\section{Conditional feature contributions (CFCs) \label{sec:TreeInterpreter}}

The conventional wisdom of estimating the impact of a feature in tree based models is to measure the \textbf{node-wise reduction of a loss function}, such as the variance of the output $Y$, and compute a weighted average of all nodes over all trees for that feature. By its definition, such a \textit{mean decrease in impurity} (MDI) serves only as a global measure and is typically not used to explain a \textit{per-observation, local impact}.
\cite{saabas2019treeInterpreter} proposed the novel idea of explaining a prediction by following the
decision path and attributing changes in the expected output of the model to each feature along the path.
\begin{figure}[!htbp]
  \centering
  \includegraphics[height=4in]{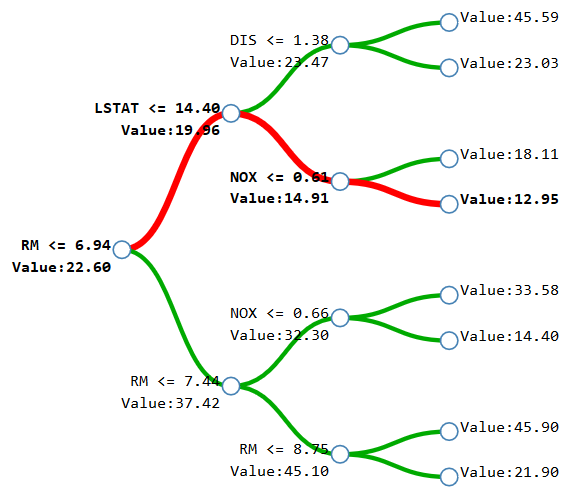}
  \caption{ Taken from \citep{saabas2019RFexplain}: Depicted is a regression decision tree to predict housing prices. The tree has conditions on each internal node and a value associated with each leaf (i.e. the value to be predicted). But additionally, the value at each internal node i.e. the mean of the response variables in that region, is shown. The red path depicts an example prediction $Y=12.95$, broken down as follows:\\  $\mathbf{12.95} \approx \mathbf{22.60} (\bar{Y}) - \mathbf{2.64} \text{(loss from RM)} - \mathbf{5.04} \text{(loss from LSTAT)} - \mathbf{1.96}\text{ (loss from NOX)}$}
  \label{fig:BostonHousing1}
\end{figure}
Figure \ref{fig:BostonHousing1} illustrates the main idea of decomposing each prediction through the sequence of regions that correspond to each node in the tree. Each decision either adds or subtracts from the value given in the parent node and can be attributed to the feature at the node. So, each individual prediction can be defined as the global mean plus the sum of the $K$ feature contributions:
 \begin{equation}
 \label{eq:TreeDecomposition}
 f_{pred}(x_i) =   \bar{Y} + \sum_{k = 1}^K{f_{T, k}(x_i)}
 \end{equation}
where $f_{T, k}(x_i)$ is the contribution from the $k$-th feature (for tree T), written as a sum over all the inner nodes $t$ such that $v(t) = k$  \citep{NIPS2019_9017}\footnote{ Appendix \ref{sec:TreeDefs} contains expanded definitions and more thorough notation.}:
 \begin{equation}
 \label{eq:fTk}
 f_{T, k}(x_i)=\sum_{t \in I(T): v(t)=k}\left[\mu_{n}\left(t^{left}\right) \mathbbm{1}\left(x_i \in R_{t^{left}}\right)+\mu_{n}\left(t^{right}\right) \mathbbm{1}\left(x_i \in R_{t^{right}}\right)-\mu_{n}(t) \mathbbm{1}\left(x_ \in R_{t}\right)\right]
 \end{equation}
 where $v(t)$ is the feature chosen for the split at node $t$.
 
A ``local'' feature importance score can be obtained by summing Eq. (\ref{eq:fTk}) over all trees. Adding these local explanations over all data points yields a ``global'' importance score: 
 \begin{equation}
 \label{eq:TreeDecomposition2}
 Imp_{global}(k) =  \sum_{i = 1}^N{|Imp_{local}(k,x_i)|} =  \sum_{i = 1}^N{ \frac{1}{n_T} \sum_{T}{|f_{T, k}(x_i)|}}
 \end{equation}
In the light of wanting to explain the predictions from tree based machine learning models,  the ``Saabas algorithm'' is extremely appealing, because
\begin{itemize}
  \item The positive and negative contributions from nodes convey directional information unlike the strictly positive purity gains.
  \item By combining many local explanations we can represent global structure while retaining local faithfulness to the original model.
  \item The expected value of every node in the tree can be estimated efficiently by averaging the model output over all the training samples that pass through that node.
  \item The algorithm has been implemented and is easily accessible in a python \citep{saabas2019treeInterpreter} and R \citep{tree.interpreter2020} library.
\end{itemize}
However, \cite{lundberg2020local} pointed out  that it is strongly biased to alter the impact of features based on their distance from the root of a tree. This causes Saabas values to be inconsistent, which means one can modify a model to make a feature clearly more important, and yet the Saabas value attributed to that feature will decrease.
As a solution, the authors developed an algorithm (``TreeExplainer'') that computes local explanations based on exact Shapley values in polynomial time. This provides local explanations with theoretical guarantees of local accuracy and consistency. A python library is available at \url{https://github.com/slundberg/shap}. One should not forget though that the same idea of adding \textit{conditional feature contributions} lies at the heart of  \textit{TreeExplainer}.

In this section, we call attention to another source of bias which is the result of using the same (inbag) data to (i) greedily split the nodes during the growth of the tree and (ii) computing the node-wise changes in prediction.
We use the well known titanic data set to illustrate the perils of putting too much faith into importance scores which are based entirely on training data - not on OOB samples - and make no attempt to discount node splits in deep trees that are spurious and will not survive in a validation set.\\
In the following model\footnote{In all random forest simulations, we choose $mtry=2, ntrees=100$ and exclude rows with missing \textit{Age}} we include \textit{passengerID} as a feature along with the more reasonable \textit{Age}, \textit{Sex} and \textit{Pclass}.

\begin{figure}[!htbp]
  \centering
  \includegraphics[width=0.85 \textwidth]{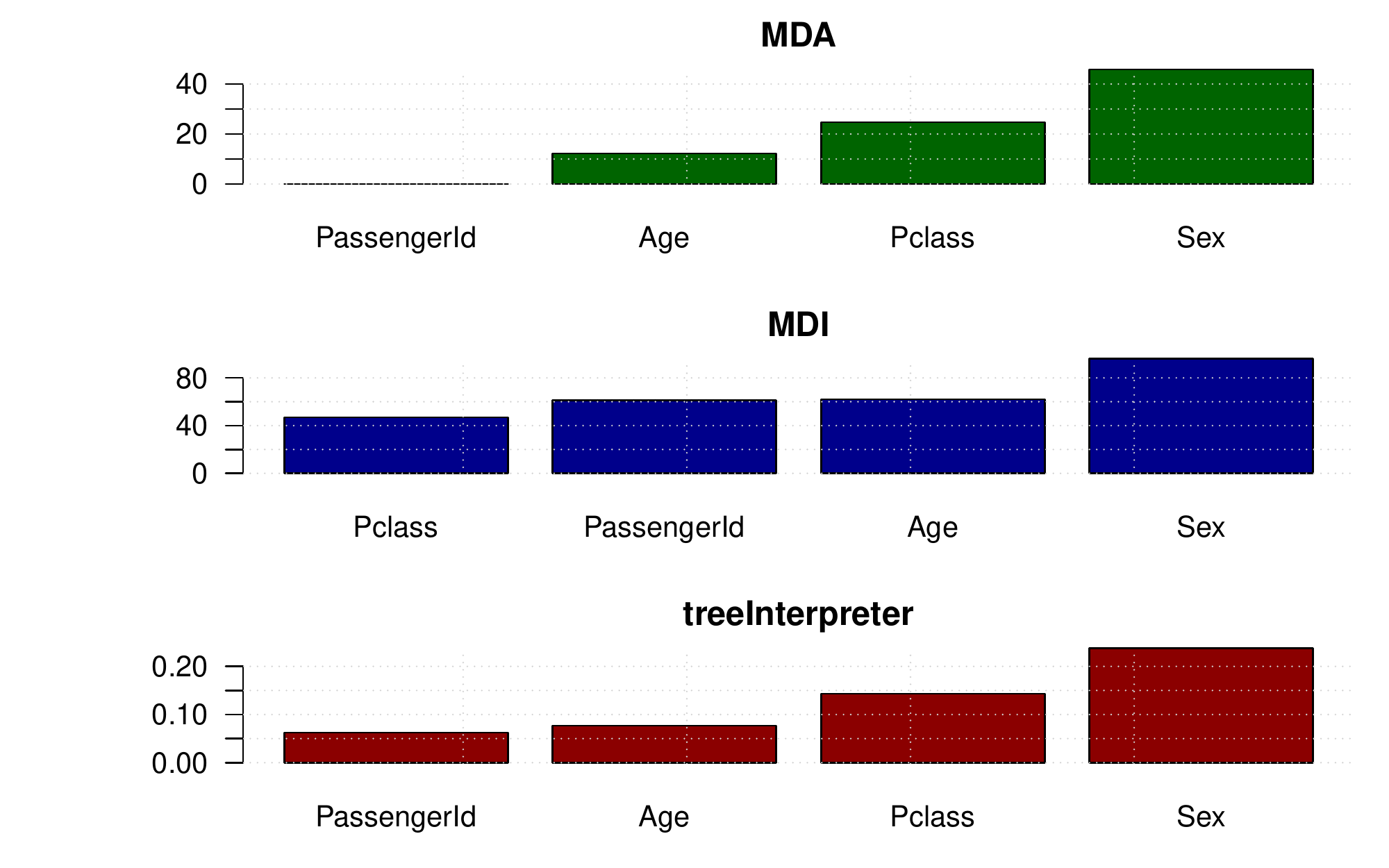}%
  \caption{Permutation importance  (MDA, top panel) versus Mean decrease impurity  (MDI, left panel) versus conditional feature contributions (TreeInterpreter) for  the Titanic data. The permutation based importance (MDA) is not fooled by the irrelevant ID feature. This is maybe not unexpected as the IDs should bear no predictive power for the out-of-bag samples.}
  \label{fig:titanic2}
\end{figure}

Figure~\ref{fig:titanic1} below depicts the distribution of the individual, ``local'' feature contributions, preserving their sign.
\begin{figure}[!htbp]
  \centering
  \includegraphics[width=0.9\textwidth]{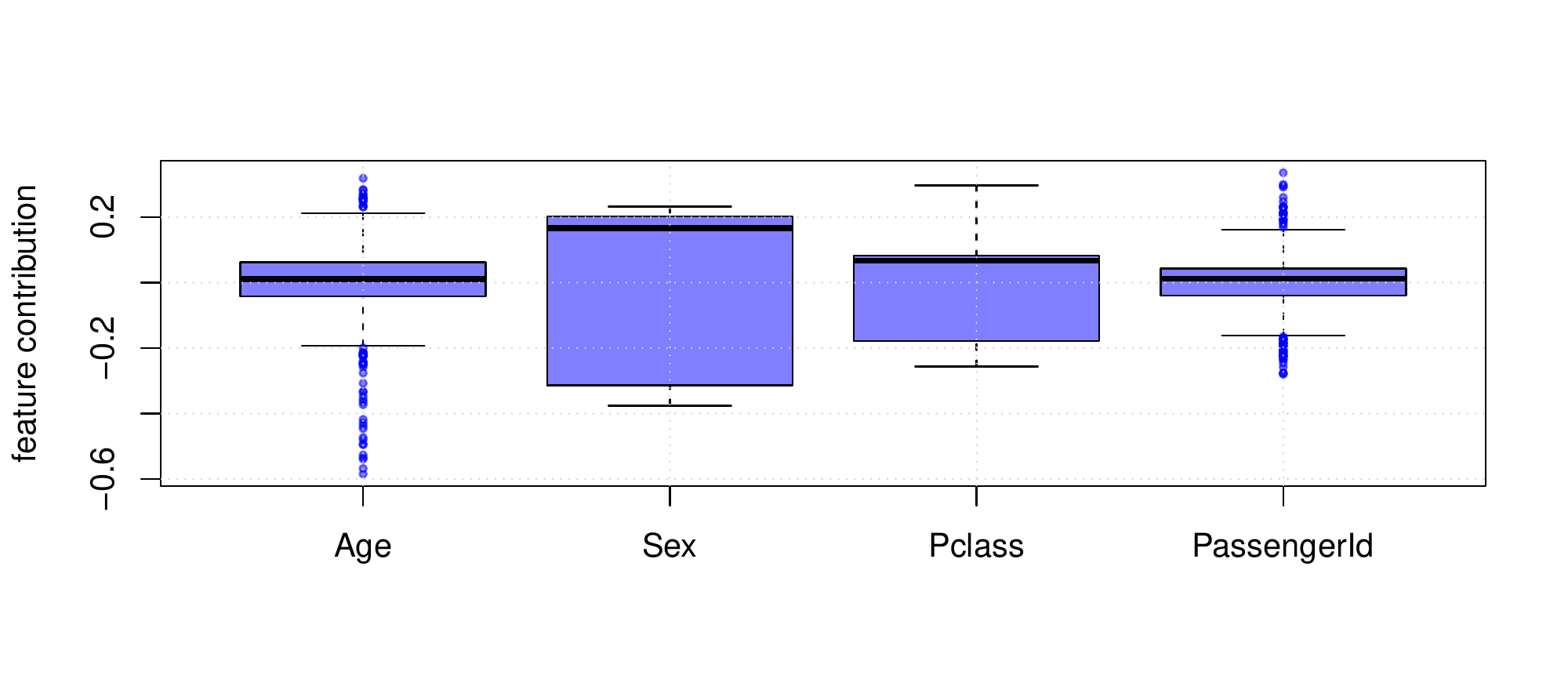}%
  \caption{Conditional feature contributions (TreeInterpreter) for  the Titanic data.}
  \label{fig:titanic1}
\end{figure}
The large variations for the variables with high cardinality (\textit{Age, passengerID}) are worrisome.
We know that the impact of \textit{Age} was modest and that \textit{passengerID} has no impact on survival but when we sum the absolute values, both features receive sizable importance scores, as shown in Figure~\ref{fig:titanic2}. This troubling result is robust to random shuffling of the ID.
Section \ref{sec:EquivMDI} will point out a close analogy between the well known MDI score and the more recent measure based on the conditional feature contributions.

\section{SHAP values \label{sec:TreeExplainer}}

\cite{lundberg2020local} introduce a new local feature attribution method for trees based on \textbf{SHapley Additive exPlanation} (SHAP) values which fall in the class of \textit{additive feature attribution methods}. The authors point to results from game theory implying that Shapley values are the only way to satisfy three important
properties: \textit{local accuracy, consistency, and missingness}.
\begin{figure}[!htb]
\minipage{0.6\textwidth}
  \includegraphics[width=\textwidth]{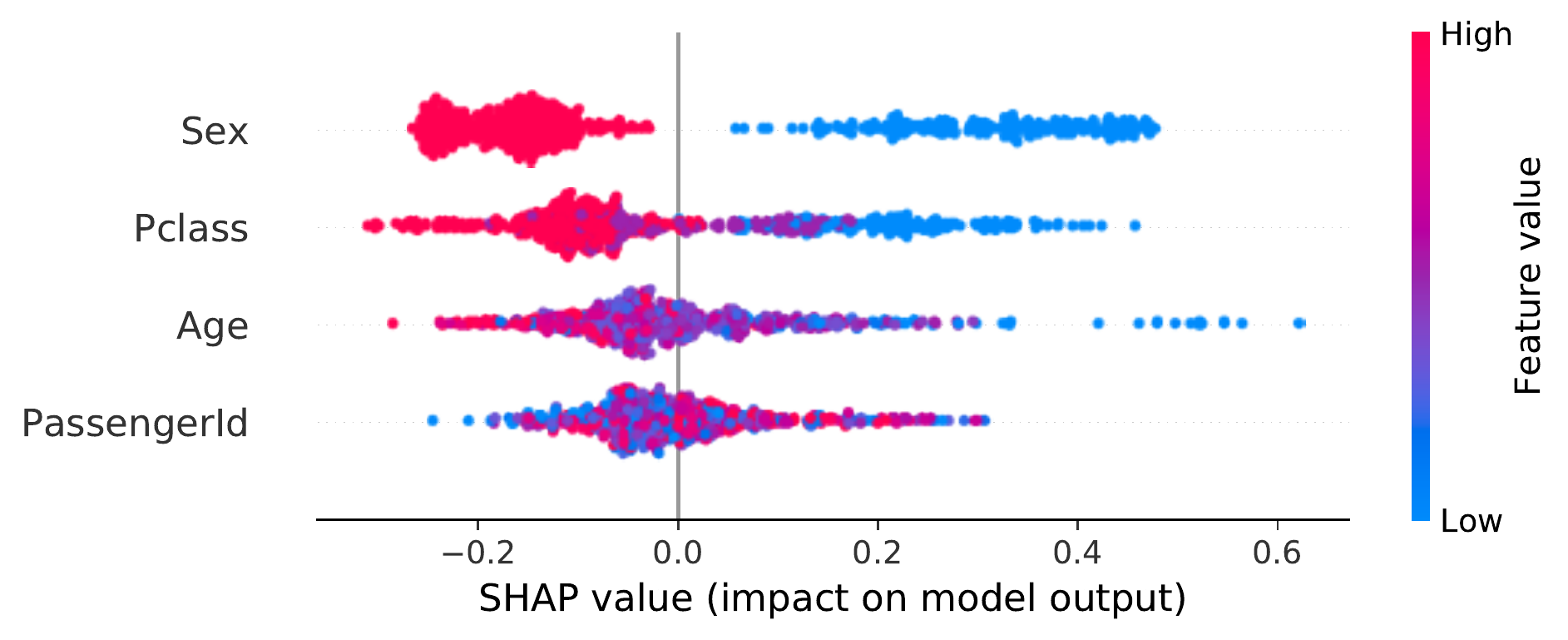}
  \endminipage\hfill
  \minipage{0.39\textwidth}
  \includegraphics[width=\textwidth]{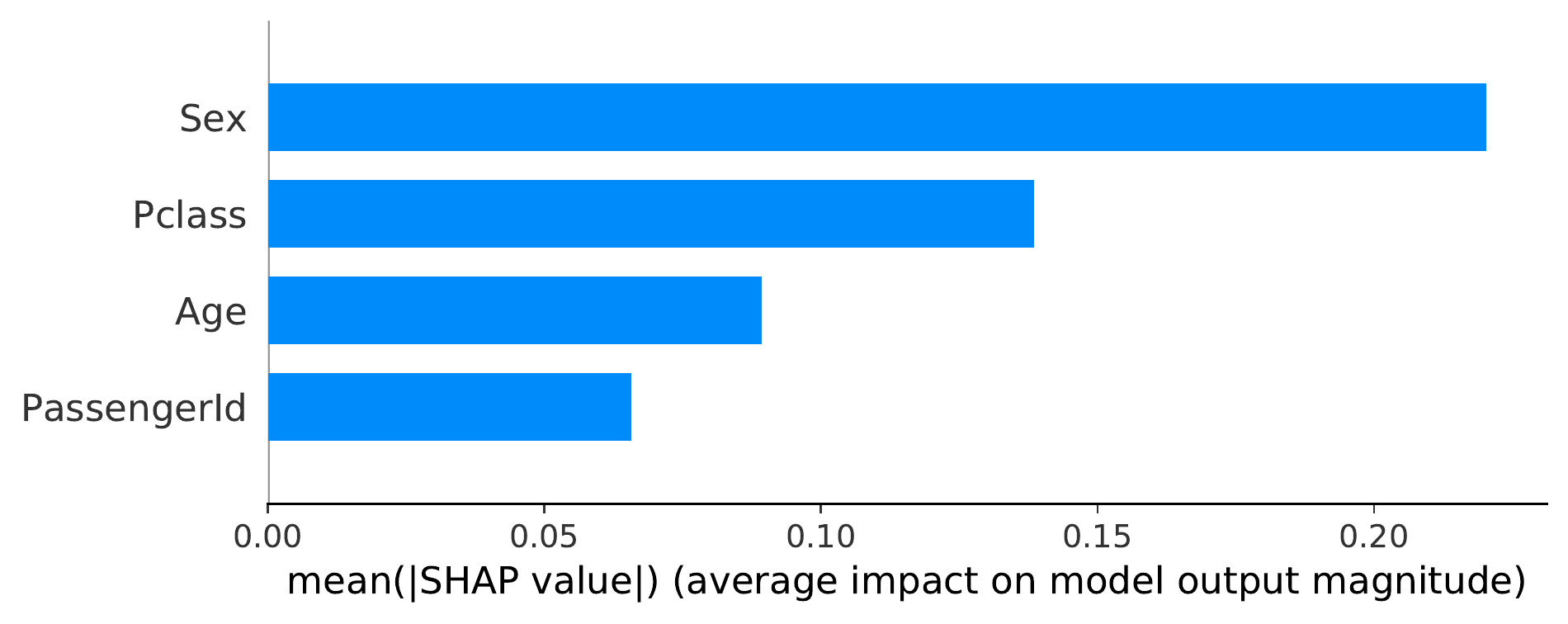}
\endminipage
 \caption{ SHapley Additive exPlanation (SHAP) values (TreeExplainer) for  the Titanic data.}
  \label{fig:titanicTreeExplainer}
\end{figure}
In this section we show that even (SHAP) values suffer from (i) a strong dependence on feature cardinality, and (ii) assign non zero importance scores to uninformative features, which would violate the \textit{missingness} property.
\begin{figure}[!htb]
  \centering
  \includegraphics[width=0.9\textwidth]{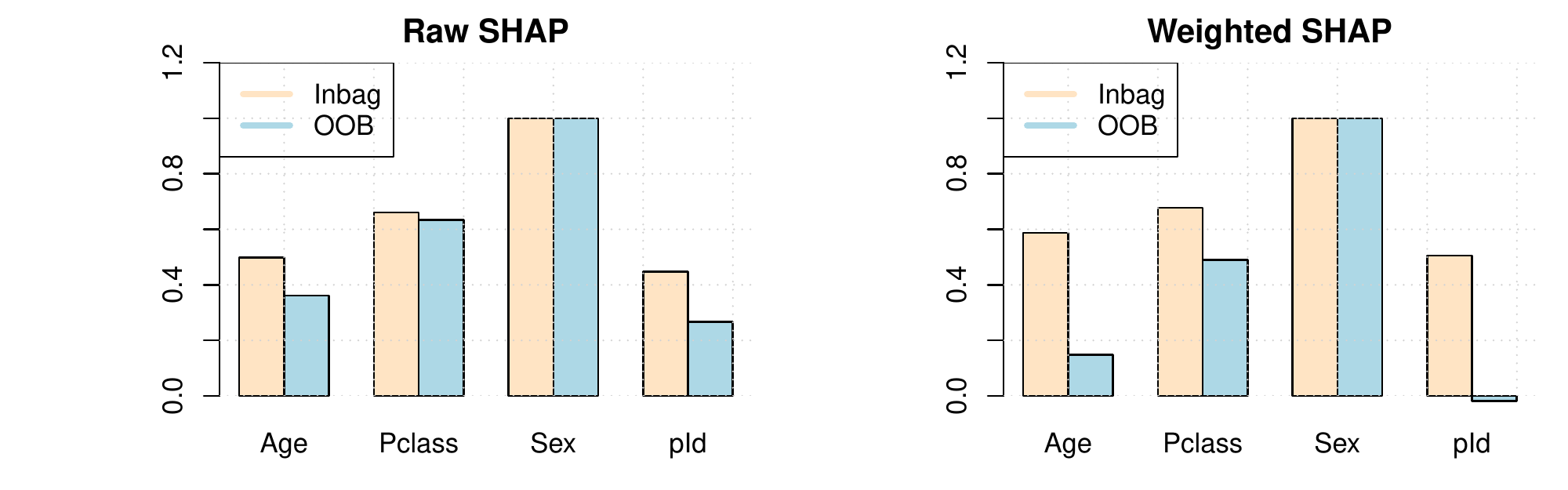}%
  \caption{Left graph: ``raw'' SHAP values for  the Titanic data, separately computed for inbag and OOB.  Right graph: weighted SHAP values are multiplied by $y_i$ before averaging which eliminates the spurious contributions due to \textit{passengerID} for OOB. Note that we scaled the SHAP values to their respective maxima for easier comparison. (\textit{pId} is short for \textit{passengerID}) }
  \label{fig:Titanic_SHAP_Wghted}
\end{figure}
We begin by extending the Titanic example from the previous section and find that \textit{TreeExplainer} also assigns a non zero value of feature importance to  \textit{passengerID}, as shown in Figure~\ref{fig:titanicTreeExplainer},  which is due to mixing inbag and out-of-bag data for the evaluation.
Simply separating the inbag from the OOB SHAP values is not a remedy as shown in the left graph of Figure~\ref{fig:Titanic_SHAP_Wghted}.
However, inspired by \cite{NIPS2019_9017} (see section \ref{sec:EquivMDI}), we compute weighted SHAP values by mutliplying with $y_i$ before averaging. The right graph of Figure~\ref{fig:Titanic_SHAP_Wghted} illustrates the elimination of the spurious contributions due to \textit{passengerID} for the OOB SHAP values.
Further support for the claims above is given by the following two kinds of simulations.

\subsection{Null/Power Simulations \label{sec:SimulatedData}}

We replicate the simulation design used by \cite{Strobl2007a} where a binary response variable Y is  predicted from a set of $5$ predictor variables that vary in their scale of measurement and
number of categories. The first predictor variable $X_1$ is continuous, while the other predictor variables $X_2 ,\ldots, X_5$ are
multinomial with $2, 4, 10, 20$ categories, respectively. 
The sample size for all simulation studies was set to n = 120.
In the first \textit{null case}  all predictor variables and the response are sampled
independently. We would hope that a reasonable variable importance measure would not prefer any one predictor variable over any other.
In the second simulation study, the so-called \textit{power case},
 the distribution of the response is a binomial process with probabilities that depend on the value
of $x_2$, namely $P(y=1|X_2=1)=0.35, P(y=1|X_2=2)=0.65$ .

\begin{figure}[htbp]
  \centering
  \includegraphics[width=\textwidth]{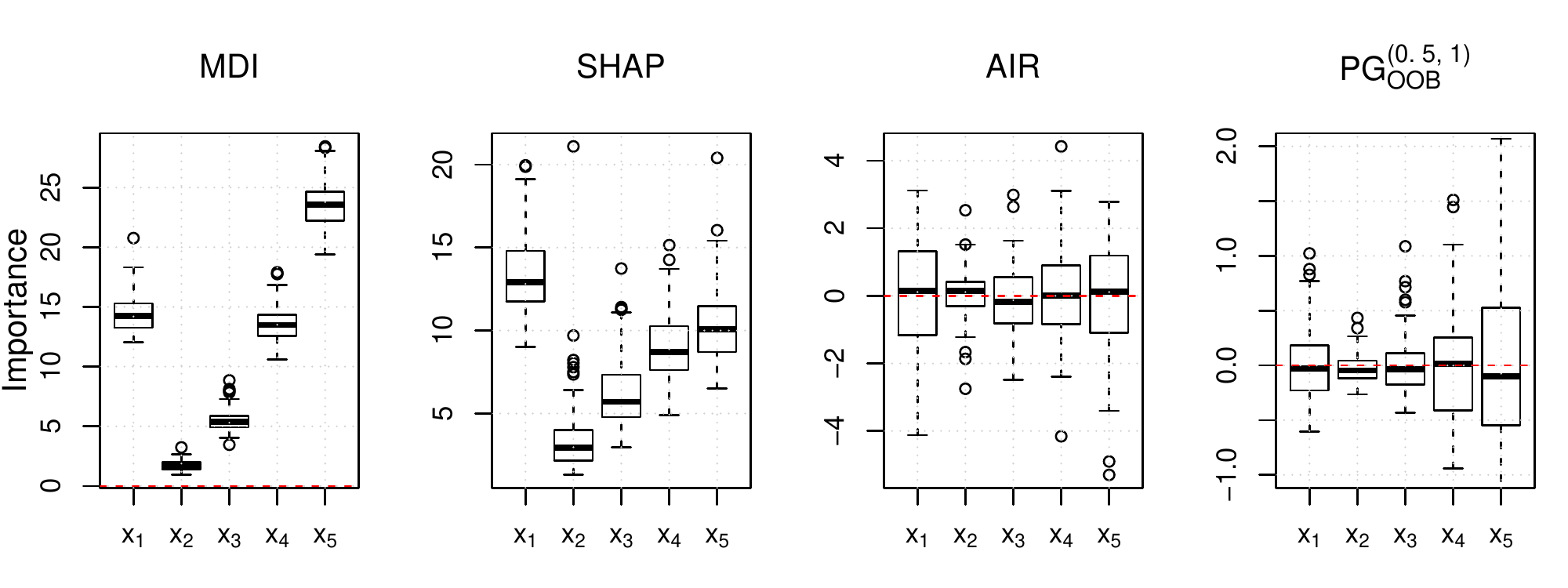}%
  \caption{Results of the null case, where none of the predictor variables is informative. }
  \label{fig:NullSim}
\end{figure}

\begin{figure}[htbp]
  \centering
  \includegraphics[width=\textwidth]{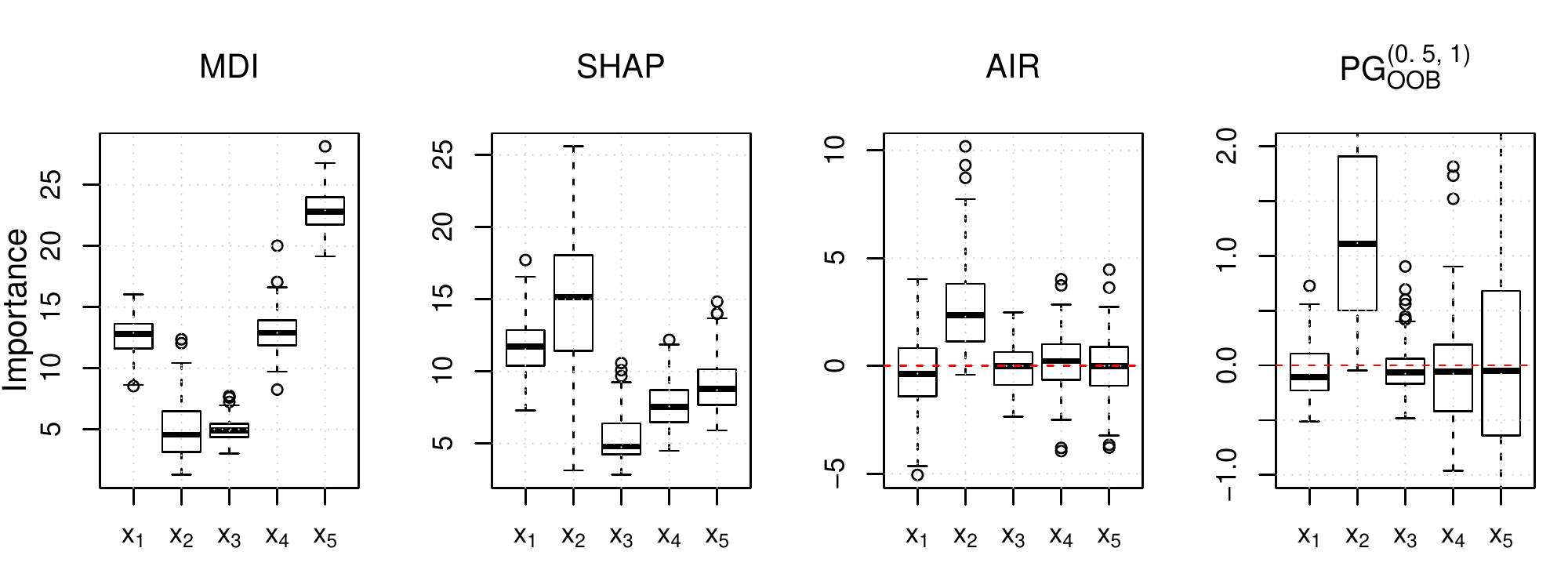}%
  \caption{Results of the power study, where only $X_2$ is informative. Other simulation details as in Fig.~\ref{fig:NullSim}. }
  \label{fig:PowerSim}
\end{figure}

As is evident in the two leftmost panels of Figure~\ref{fig:NullSim}, both the Gini importance (MDI) and the SHAP values show a strong preference for variables with many categories and the continuous variable. This bias is of course well-known for MDI but maybe unexpected for the SHAP scores which clearly violate the \textit{missingness} property.
Encouragingly, both $PG_{OOB}^{(0.5,1)}$ and AIR \citep{nembrini2018revival} yield low scores for all predictors. 
The notable differences in the variance of the distributions for predictor variables with different scale of measurement or number of categories are unfortunate but to be expected. (The larger the numbers of categories in a multinomial variable, the fewer the numbers of observations per category and the larger therefore the variability of the measured qualities of the splits performed using the multinomial variable)
The results from the power study are summarized in Figure~\ref{fig:PowerSim}.  
MDI and SHAP again show a strong bias towards variables with many
categories and the continuous variable. At the chosen sigal-to-noise ratio MDI fails entirely to identify the relevant predictor variable. In fact, the mean value for the relevant variable $X_2$ is lowest and only slightly higher than in the null case.
Both $PG_{OOB}^{(0.5,1)}$ and AIR clearly succeed in identifying $X_2$ as the most relevant feature.
The large fluctuations of the importance scores for $X_4$ and especially $X_5$ are bound to yield moderate ``false positive'' rates and incorrect rankings in single trials.
\begin{figure}[htbp]
  \centering
  \includegraphics[width=\textwidth]{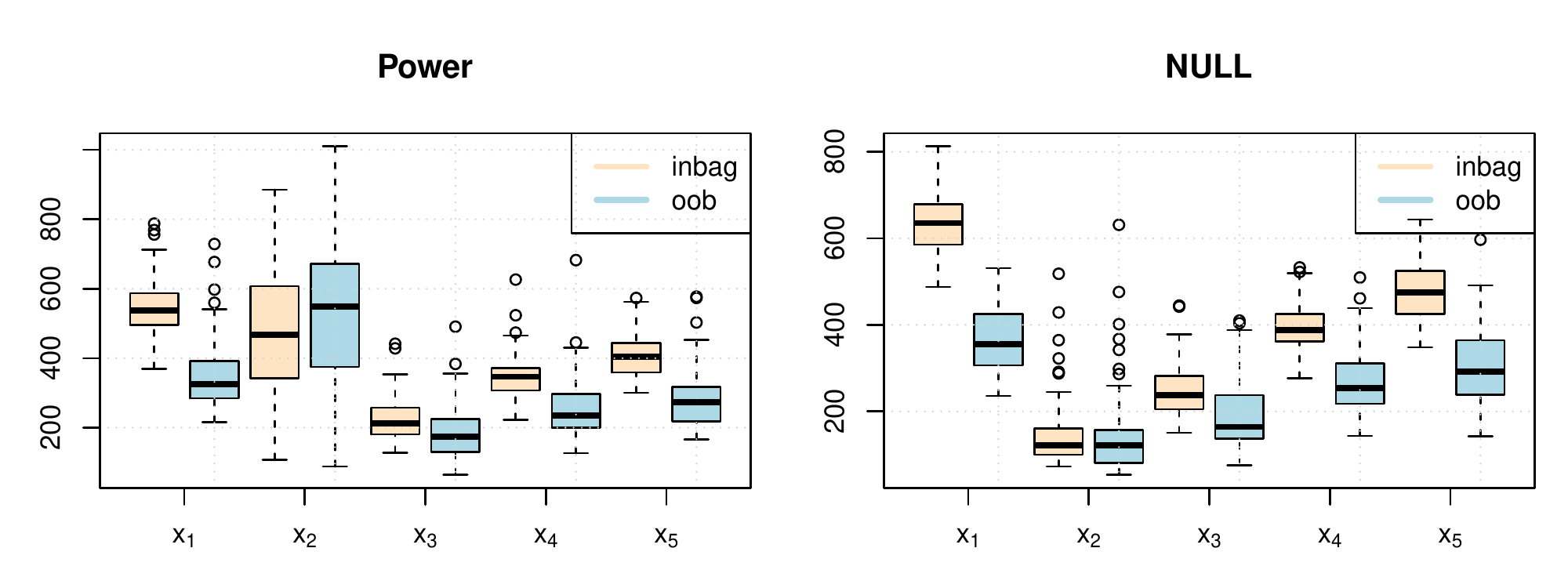}%
  \caption{Weighted SHAP values, as explined in the text. Left graph: power study, where only $X_2$ is informative. Right graph: null case, where none of the predictor variables is informative. Other simulation details as in Figs.~\ref{fig:NullSim}, \ref{fig:PowerSim}}
  \label{fig:shap_wghted}
\end{figure}
The signal-to-noise separation for the SHAP values is moderate but can be greatly improved by mutliplying with $y_i$ before averaging (in analogy to Figure~\ref{fig:Titanic_SHAP_Wghted} and section \ref{sec:EquivMDI}), as shown in Fig.~\ref{fig:shap_wghted}.

\subsection{Noisy feature identification}

For a more systematic comparison of the 4 proposed penalized Gini scores, we closely follow the simulations outlined in \cite{li2019debiased} involving discrete features with different number of distinct
values, which poses a critical challenge for MDI. The data has $1000$ samples with $50$ features. All
features are discrete, with the $j$th feature containing $j + 1$ distinct values $0,1, \ldots ,j$. We randomly
select a set $S$ of $5$ features from the first ten as relevant features. The remaining features are noisy
features. All samples are i.i.d. and all features are independent. We generate the outcomes using the
following rule: 
\[
P(Y = 1| X) = \text{Logistic}(\frac{2}{5} \sum_{j \in S}{x_j/j} -1)
\]
Treating the noisy features as label $0$ and the relevant features as label $1$, we can evaluate a feature
importance measure in terms of its area under the receiver operating characteristic curve (AUC). 
\begin{table}[ht]
\centering
\begin{tabular}{|cc|cc|ccc|ccc|}
  \hline
 $\widehat{PG}_{oob}^{(1,0)}$ & $PG_{oob}^{(1,0)}$ & $\widehat{PG}_{oob}^{(0.5,1)}$ & $PG_{oob}^{(0.5,1)}$ & SHAP & SHAP$_{in}$ & SHAP$_{oob}$ & AIR & MDA & MDI \\ 
  \hline
 {\color{blue} 0.66} & 0.28 & {\color{blue} 0.92} & 0.78 & 0.66 & 0.56 & 0.73 & 0.68 & 0.65 & 0.10 \\ 
   \hline
\end{tabular}
\caption{Average AUC scores for noisy feature identification. \textit{MDA} = permutation importance, \textit{MDI} = (default) Gini impurity. The $\widehat{PG}_{oob}$ scores apply the variance bias correction $n/(n-1)$. The SHAP$_{in}$, SHAP$_{oob}$ scores are based upon separating the inbag from the oob data.}
  \label{tbl:NoisyFtrSim}
\end{table}
We grow $100$ deep trees (minimum leaf size equals 1, $m_{try}=3$), repeat the whole process $100$ times and report the average AUC scores for each method in
Table \ref{tbl:NoisyFtrSim}. For this simulated setting, $\widehat{PG}_{oob}^{(0.5,1)}$ achieves the best AUC score under all cases, most likely because of the separation of the signal from noise mentioned above. We notice that the AUC score for the OOB-only $\widehat{PG}_{oob}^{(1,0)}$ is competitive to the permutation importance, SHAP and the AIR score.

\section{MDI versus CFCs \label{sec:EquivMDI}}

As elegantly demonstrated by \cite{NIPS2019_9017}, the MDI of  feature $k$ in a tree $T$ can be written as
 \begin{equation}
 \label{eq:MDIDecomposition}
 MDI =    \frac{1}{\left|\mathcal{D}^{(T)}\right|} \sum_{i \in \mathcal{D}^{(T)}}{f_{T, k}(x_i) \cdot y_i}
 \end{equation}
where $\mathcal{D}^{(T)}$ is the bootstrapped or subsampled data set of the original data $\mathcal{D}$.
Since $\sum_{i \in \mathcal{D}^{(T)}}{f_{T, k}(x_i) } = 0$, we can view MDI  essentially as 
the sample covariance between $f_{T, k}(x_i)$ and $y_i$ on the bootstrapped dataset $\mathcal{D}^{(T)}$.
\begin{figure}[!htbp]
  \centering
  \includegraphics[width=\textwidth]{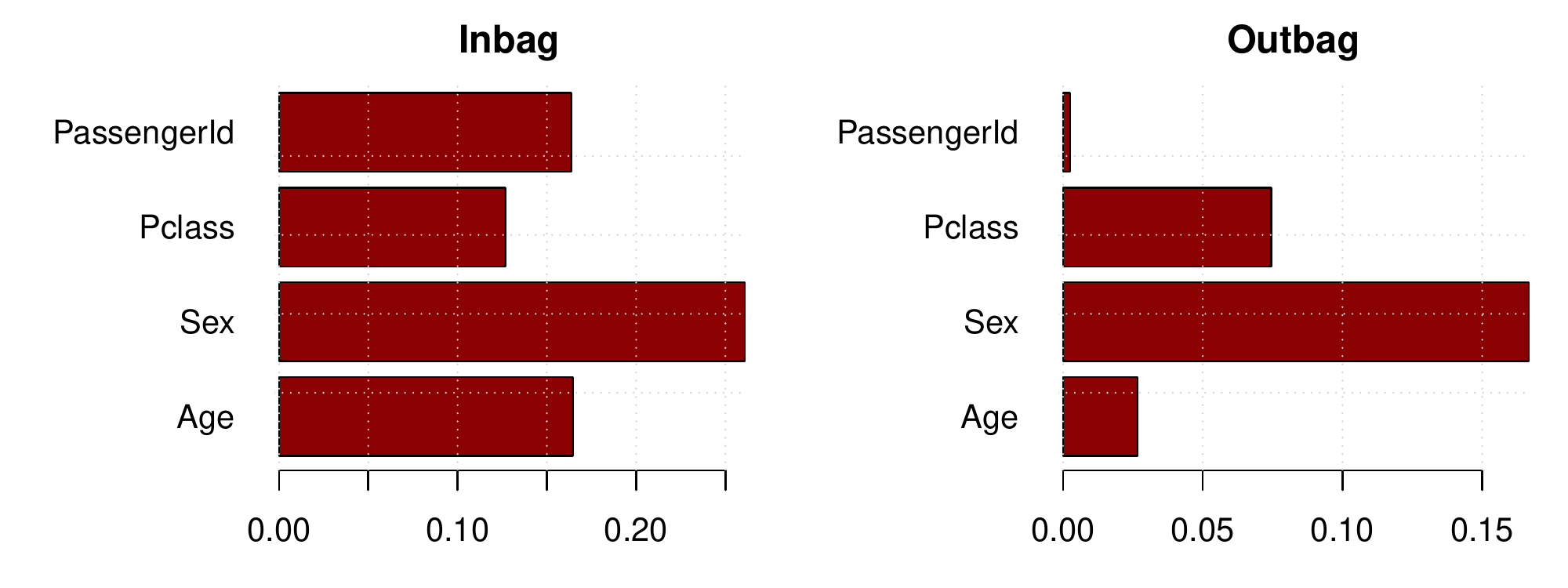}%
  \caption{MDI as a restricted sum of conditional feature contributions defined by Eq. (\ref{eq:MDIDecompositionBinary}) for the (left panel) inbag and (right panel) outbag data, respectively.}
  \label{fig:titanic3}
\end{figure}
Alternatively, we can view MDI as a particular weighted average of the CFCs, which for the special case of binary classification ($y_i \in \{0,1\}$) means that one only adds up those CFCs for which $y_i = 1$.
 \begin{equation}
 \label{eq:MDIDecompositionBinary}
 MDI =    \frac{1}{\left|\mathcal{D}^{(T)}[y_i = 1]\right|} \sum_{i \in \mathcal{D}^{(T)}[y_i = 1]}{f_{T, k}(x_i)}
 \end{equation}
Figure~\ref{fig:titanic3} shows this new measure separately for the inbag and outbag components; the middle panel  of Figure \ref{fig:titanic2} is proportional to the left panel.
The misleadingly high contribution of \textit{passengerID} is due to the well known shortcoming of MDI: RFs use the training data $\mathcal{D}^{(T)}$ to construct the functions $f_{T, k}()$ , then MDI uses the same data to evaluate (\ref{eq:MDIDecomposition}).

\section{Debiasing MDI via oob samples \label{sec:EquivDebiased}}

In this section we give a short version of the proof that $PG_{oob}^{(1,2)}$ is equivalent to the \textit{MDI-oob} measure defined in \cite{NIPS2019_9017}. For clarity we assume binary classification; Appendix \ref{sec:EquivDebiased} contains an expanded version of the proof including the multi-class case. 
\textit{MDI-oob}  is based on the usual variance reduction per node as shown in Eq. (34) (proof of Proposition (1)), but with a ``variance'' defined as the mean squared deviations of $y_{oob}$ from the inbag mean $\mu_{in}$:
\[
 \Delta_{I}(t) = \frac{1}{N_n(t)} \cdot  \sum_{i \in D(T)}{(y_{i,oob} - \mu_{n,in})^2}  1 (x_i \in R_t) - \ldots
\]
We can, of course, rewrite the variance as 
\begin{align}
\frac{1}{N_n(t)} \cdot \sum_{i \in D(T)}{(y_{i,oob} - \mu_{n,in})^2}  & = & \frac{1}{N_n(t)} \cdot  \sum_{i \in D(T)}{(y_{i,oob} - \mu_{n,oob})^2} & + (\mu_{n,in} - \mu_{n,oob})^2 \\
 & = & p_{oob} \cdot (1- p_{oob} ) & + (p_{in} - p_{oob})^2 
\end{align}
where the last equality is for Bernoulli $y_i$, in which case the means $\mu_{in/oob}$ become proportions $p_{in/oob}$ and the first sum is equal to the binomial variance $p_{oob} \cdot (1- p_{oob} )$.
The final expression is effectively equal to $PG_{oob}^{(1,2)}$.

Lastly, we now show that $PG_{oob}^{(0.5,1)}$ is equivalent to the \textit{unbiased split-improvement} measure defined in \cite{zhou2019unbiased}.
For the binary classificaton case, we can rewrite $PG_{oob}^{(0.5,1)}$ as follows:
\begin{align}
PG_{oob}^{(0.5,1)} &=&  \frac{1}{2} \cdot \sum_{d=1}^D{ \hat{p}_{d,oob} \cdot \left(1- \hat{p}_{d,oob} \right)    +  \hat{p}_{d,in} \cdot \left(1- \hat{p}_{d,in} \right)    + (\hat{p}_{d,oob} - \hat{p}_{d,in})^2} \\
 &=& \hat{p}_{oob} \cdot \left(1- \hat{p}_{oob} \right) +  \hat{p}_{in} \cdot \left(1- \hat{p}_{in} \right)    + (\hat{p}_{oob} - \hat{p}_{in})^2 \\
 &=& \hat{p}_{oob} - \hat{p}_{oob}^2 +  \hat{p}_{in} - \hat{p}_{in}^2    + \hat{p}_{oob}^2 - 2  \hat{p}_{oob} \cdot \hat{p}_{in} +  \hat{p}_{in}^2\\
&=&  \hat{p}_{oob}   +   \hat{p}_{in} - 2  \hat{p}_{oob} \cdot  \hat{p}_{in} 
\end{align}


\section{Discussion}

Random forests and gradient boosted trees are among the most popular and powerful \citep{olson2017data} non-linear predictive models used in a wide variety of fields.
\cite{lundberg2020local} demonstrate that tree-based models can be more accurate than neural networks and even more interpretable than linear models.
In the comprehensive overview of variable importance in regression models \cite{gromping2015variable} distinguishes between methods based on (i) variance decomposition and (ii) standardized coefficient sizes, which is somewhat analogous to the difference between (i) MDI and (ii) CFCs. The latter measure the directional impact of $x_{k,i}$ on the outcome $y_i$, whereas MDI based scores measure a kind of \textit{partial} $R^2_{k}$ (if one stretched the analogy to linear models).
\cite{NIPS2019_9017} ingeniously illustrate the connection between these seemingly fundamentally different methods via eq. (\ref{eq:MDIDecomposition}).
And the brilliant extension of CFCs to their Shapley equivalents by \cite{lundberg2020local}
bears affinity to the game-theory-based metrics LMG and PMVD (\cite{gromping2015variable} and references therein), which are based on averaging the sequential $R^2_{k}$ over all orderings of regressors.

In this paper we have (i) connected the proposals to reduce the well known bias in MDI by mixing inbag and oob data \citep{NIPS2019_9017,zhou2019unbiased} to a common framework \citep{loecher2020unbiased}, and (ii) pointed out that similar ideas would benefit/debias the \textit{conditional feature contributions} (CFCs)  \citep{saabas2019treeInterpreter} as well as the related  \textit{SHapley Additive exPlanation} (SHAP) values \citep{lundberg2020local}.
We mention in passing that \citep{loecher2021data} demonstrated very high correlations between SHAP and CFC values for a highly diverse collection of classification data sets.

While the main findings are applicable to any tree based method, they are most relevant to random forests (RFs) since (i) oob data are readily available and (ii) RFs typically grow deep trees. \cite{NIPS2019_9017} showed a strong dependence of the MDI bias on the depth of the tree: splits in nodes closer to the roots are much more stable and supported by larger sample sizes and hence hardly susceptible to bias.  
RFs ``get away'' with the individual overfitting of deep trees to the training data by \textbf{averaging} many (hundreds) of separately grown deep trees and often achieve a favorable balance in the bias-variance tradeoff. One reason is certainly that the noisy predictions from individual trees ``average out'', which is not the case for the summing/averaging of the strictly positive MDI leading to what could be called \textit{interpretational overfitting}.
The big advantage of conditional feature contributions is that positive and negative contributions can cancel across trees making it less prone to that type of overfitting.
However, we have provided evidence that both the CFCs as well as the SHAP values are still susceptible to ``overfitting'' to the training data and can benefit from evaluation on oob data.





\section{Appendix}

\subsection{Background and notations\label{sec:TreeDefs}}

Definitions needed to understand Eq. (\ref{eq:fTk}).
(The following paragraph closely follows the definitions in \cite{NIPS2019_9017}.)

Random Forests (RFs) are an ensemble of classification and regression trees, where each tree $T$ defines a mapping from the feature space to the response. Trees are constructed independently of one another on a bootstrapped or subsampled data set $\mathcal{D}^{(T)}$ of the original data $\mathcal{D}$. Any node $t$ in a tree $T$ represents a subset (usually a hyper-rectangle) $R_{t}$ of the feature space. A split of the node $t$ is a pair $(k, z)$ which divides the hyper-rectangle $R_{t}$ into two hyper-rectangles $R_{t} \cap \mathbbm{1}\left(X_{k} \leq z\right)$ and $R_{t} \cap \mathbbm{1}\left(X_{k}>z\right)$ corresponding to the left child $t$ left and right child $t$ right of node $t$, respectively. For a node $t$ in a tree
$T, N_{n}(t)=\left|\left\{i \in \mathcal{D}^{(T)}: \mathbf{x}_{i} \in R_{t}\right\}\right|$ denotes the number of samples falling into $R_{t}$ and
$$
\mu_{n}(t):=\frac{1}{N_{n}(t)} \sum_{i: \mathbf{x}_{i} \in R_{t}} y_{i}
$$
 We define the set of inner nodes of a tree $T$ as $I(T)$.

\subsection{Variance Reduction View \label{sec:ProofDebiased}}

Here, we provide a full version of the proof sketched in section \ref{sec:EquivDebiased} which leans heavily on the proof of Proposition (1) in \cite{li2019debiased} .

We consider the usual variance reduction per node but with a ``variance'' defined as the mean squared deviations of $y_{oob}$ from the inbag mean $\mu_{in}$:

\begin{equation}\label{eq:Li33}
\begin{split} 
 \Delta_{\mathcal{I}}(t) &=\frac{1}{N_{n}(t)} \sum_{i \in \mathcal{D}^{(T)}}\left[y_{i,oob}-\mu_{n, in}(t)\right]^{2} \mathbbm{1}\left(\mathbf{x}_{i} \in R_{t}\right) \\ &-\left[y_{i,oob}-\mu_{n, in}\left(t^{\text {left}}\right)\right]^{2} \mathbbm{1}\left(\mathbf{x}_{i} \in R_{t^{\text {left}}}\right)-\left[y_{i,oob}-\mu_{n, in}\left(t^{\text{right}}\right)\right]^{2} \mathbbm{1}\left(\mathbf{x}_{i} \in R_{\text{right}}\right) 
\end{split}
\end{equation}

\begin{equation}\label{eq:VarSplit1}
\begin{split} 
  =\frac{1}{N_{n}(t)} \sum_{i \in \mathcal{D}^{(T)}} & \left( \left[y_{i,oob}-\mu_{n, oob}(t)\right]^{2} + \left[\mu_{n,in}(t) - \mu_{n,oob}(t) \right]^2 \right) \mathbbm{1}\left(\mathbf{x}_{i} \in R_{t}\right) \\ 
  &- \left( \left[y_{i,oob}-\mu_{n, oob}(t^{\text {left}})\right]^{2} + \left[\mu_{n,in}(t^{\text {left}}) - \mu_{n,oob}(t^{\text {left}}) \right]^2 \right)  \mathbbm{1}\left(\mathbf{x}_{i} \in R_{t^{\text {left}}}\right) \\
  & - \left( \left[y_{i,oob}-\mu_{n, oob}(t^{\text {right}})\right]^{2} + \left[\mu_{n,in}(t^{\text {right}}) - \mu_{n,oob}(t^{\text {right}}) \right]^2 \right) \mathbbm{1}\left(\mathbf{x}_{i} \in R_{\text{right}}\right) 
\end{split}
\end{equation}

  \begin{align*}
\label{eq:VarSplit2}
   &=\frac{1}{N_{n}(t)} \underbrace{ \sum_{i \in \mathcal{D}^{(T)}}{  \left\{  \left[y_{i,oob}-\mu_{n, oob}(t)\right]^{2}\mathbbm{1}\left(\mathbf{x}_{i} \in R_{t}\right) \right\}}}_{N_{n}(t) \cdot p_{oob}(t) \cdot (1- p_{oob}(t) )} + \underbrace{\left[\mu_{n,in}(t) - \mu_{n,oob}(t) \right]^2}_{\left[p_{oob}(t) - p_{in}(t)\right]^2} \\
  &- \frac{1}{N_{n}(t)} \underbrace{ \sum_{i \in \mathcal{D}^{(T)}}  \left\{ \left[y_{i,oob}-\mu_{n, oob}(t^{\text {left}})\right]^{2}  \mathbbm{1}\left(\mathbf{x}_{i} \in R_{t^{\text {left}}}\right) \right\}}_{N_{n}(t^{\text {left}}) \cdot p_{oob}(t^{\text {left}}) \cdot (1- p_{oob}(t^{\text {left}}) )} + \underbrace{ \left[\mu_{n,in}(t^{\text{left}}) - \mu_{n,oob}(t^{\text {left}}) \right]^2}_{\left[p_{oob}(t^{\text {left}}) - p_{in}(t^{\text {left}})\right]^2} \\
  & - \frac{1}{N_{n}(t)} \underbrace{  \sum_{i \in \mathcal{D}^{(T)}}  \left\{ \left[y_{i,oob}-\mu_{n, oob}(t^{\text {right}})\right]^{2} \mathbbm{1}\left(\mathbf{x}_{i} \in R_{\text{right}}\right) \right\}}_{N_{n}(t^{\text {right}}) \cdot p_{oob}(t^{\text {right}}) \cdot (1- p_{oob}(t^{\text {right}}) )} + \underbrace{ \left[\mu_{n,in}(t^{\text{right}}) - \mu_{n,oob}(t^{\text {right}}) \right]^2}_{\left[p_{oob}(t^{\text {right}}) - p_{in}(t^{\text {right}})\right]^2}
\end{align*}
where the last equality is for Bernoulli $y_i$, in which case the means $\mu_{in/oob}$ become proportions $p_{in/oob}$ and we replace the squared deviations with the binomial variance $p_{oob} \cdot (1- p_{oob} )$.
The final expression is then
\begin{equation}\label{eq:VarSplit3}
\begin{split} 
 \Delta_{\mathcal{I}}(t) &=p_{oob}(t) \cdot \left(1- p_{oob}(t) \right) + \left[p_{oob}(t) - p_{in}(t)\right]^2\\ 
 & - \frac{N_{n}(t^{\text {left}})}{N_{n}(t)}  \left(p_{oob}(t^{\text{left}}) \cdot \left(1- p_{oob}(t^{\text{left}}) \right) + \left[p_{oob}(t^{\text{left}}) - p_{in}(t^{\text{left}})\right]^2 \right) \\
 & - \frac{N_{n}(t^{\text {right}})}{N_{n}(t)}  \left(p_{oob}(t^{\text{right}}) \cdot \left(1- p_{oob}(t^{\text{right}}) \right) + \left[p_{oob}(t^{\text{right}}) - p_{in}(t^{\text{right}})\right]^2 \right)
\end{split}
\end{equation}
which, of course, is exactly the impurity reduction due to $PG_{oob}^{(1,2)}$:
\begin{equation}\label{eq:VarSplit4}
 \Delta_{\mathcal{I}}(t) =PG_{oob}^{(1,2)}(t) - \frac{N_{n}(t^{\text {left}})}{N_{n}(t)} PG_{oob}^{(1,2)}(t^{\text{left}})   - \frac{N_{n}(t^{\text {right}})}{N_{n}(t)} PG_{oob}^{(1,2)}(t^{\text{right}}) 
\end{equation}

Another, somewhat surprising view of MDI is given by Eqs. (\ref{eq:MDIDecomposition}) and (\ref{eq:fTk}), which for binary classification reads as:
\begin{equation}
 \label{eq:fTk2}
 \begin{split} 
MDI & = \frac{1}{\left|\mathcal{D}^{(T)}\right|} \sum_{t \in I(T): v(t)=k} \sum_{i \in \mathcal{D}^{(T)}}{ \left[\mu_{n}\left(t^{left}\right) \mathbbm{1}\left(x_i \in R_{t^{left}}\right)+\mu_{n}\left(t^{right}\right) \mathbbm{1}\left(x_i \in R_{t^{right}}\right)-\mu_{n}(t) \mathbbm{1}\left(x_ \in R_{t}\right)\right] \cdot y_i} \\
& = \frac{1}{\left|\mathcal{D}^{(T)}\right|} \sum_{t \in I(T): v(t)=k}{ -  p_{in}(t)^2 + \frac{N_{n}(t^{\text {left}})}{N_{n}(t)}  p_{in}(t^{\text{left}})^2   + \frac{N_{n}(t^{\text {right}})}{N_{n}(t)}  p_{in}(t^{\text{right}})^2 }
\end{split} 
 \end{equation}
and for the oob version:
\begin{equation}
 \label{eq:fTk3}
 MDI_{oob} = -  p_{in}(t) \cdot p_{oob}(t) + \frac{N_{n}(t^{\text {left}})}{N_{n}(t)}  p_{in}(t^{\text{left}}) \cdot p_{oob}(t^{\text{left}})    + \frac{N_{n}(t^{\text {right}})}{N_{n}(t)}  p_{in}(t^{\text{right}}) \cdot p_{oob}(t^{\text{right}})
 \end{equation} 
The above expressions suggest that node impurity could be simply measured by $ - p_{in}(t)^2$, and $- p_{in}(t) \cdot p_{oob}(t)$, respectively. While this would be

\subsection{$\mathbf{E(\Delta  \widehat{PG}_{oob}^{(0)}) = 0}$}

\noindent
The  decrease in impurity ($\Delta G$) for a parent node $m$ is the weighted difference between the Gini importance\footnote{For easier notation we have (i) left the multiplier $2$ and (ii) omitted an index for the class membership}  $G(m) = \hat{p}_m  (1- \hat{p}_m )$ and those of its left and right children:
\[
\Delta G(m) =  G(m) - \left[ N_{m_l} G(m_l) - N_{m_r} G(m_r) \right] / N_m 
\]
 
\noindent
We assume that the node $m$ splits on an \textbf{uninformative} variable $X_j$, i.e.  $X_j$ and $Y$ are independent.\\
We will use the short notation $\sigma^2_{m, .} \equiv p_{m,.} (1-p_{m,.})$ for $.$ either equal to $oob$ or $in$ and rely on the following facts and notation:
\begin{enumerate}
  \item $E[\hat{p}_{m, oob}] = p_{m,oob}$ is the ``population'' proportion of the class label in the OOB test data (of node $m$).
  \item $E[\hat{p}_{m, in}] = p_{m,in}$ is the ``population'' proportion of the class label in the inbag test data (of node $m$).
  \item $E[\hat{p}_{m, oob}] = E[\hat{p}_{m_l, oob}] = E[\hat{p}_{m_r, oob}] =p_{m,oob}$
  \item  $E[\hat{p}_{m, oob}^2]  = var(\hat{p}_{m, oob}) + E[\hat{p}_{m, oob}]^2 = \sigma^2_{m, oob}/N_m + p_{m,oob}^2$ \\
$\Rightarrow E[G_{oob}(m)]  = E[\hat{p}_{m, oob}] - E[\hat{p}_{m, oob}^2] =  \sigma^2_{m, oob} \cdot \left(1- \frac{1}{N_m}\right)$ \\
$\Rightarrow E[\widehat{G}_{oob}(m)]  =  \sigma^2_{m, oob}$
 \item $E[\hat{p}_{m, oob} \cdot \hat{p}_{m, in}] = E[\hat{p}_{m, oob}] \cdot E[\hat{p}_{m, in}] = p_{m,oob} \cdot p_{m,in}$
\end{enumerate}
Equalities 3 and 5 hold because of the independence of the inbag and out-of-bag data as well as the independence of $X_j$ and $Y$.\\

We now show that $\mathbf{E(\Delta  PG_{oob}^{(0)}) \neq 0}$
We use the shorter notation $G_{oob} = PG_{oob}^{(0)}$:
\begin{align*}
E[\Delta G_{oob}(m)] &= E[G_{oob}(m)] - \frac{N_{m_l}}{N_{m}} E[G_{oob}(m_l)] - \frac{N_{m_r}}{N_{m}} E[G_{oob}(m_r)]   \\
& = \sigma^2_{m,oob} \cdot  \left[  1- \frac{1}{N_m} - \frac{N_{m_l}}{N_{m}} \left(1- \frac{1}{N_{m_l}}\right) - \frac{N_{m_r}}{N_{m}} \left(1- \frac{1}{N_{m_r}}\right) \right]  \\
& = \sigma^2_{m,oob} \cdot  \left[  1- \frac{1}{N_m} - \frac{N_{m_l} + N_{m_r}}{N_{m}} + \frac{2}{N_m} \right]  =  \frac{\sigma^2_{m,oob}}{N_m}
\end{align*}
 We see that there is a bias if we used only OOB data, which becomes more pronounced for nodes with smaller sample sizes. This is relevant because visualizations of random forests show that the splitting on uninformative variables happens most frequently for ``deeper'' nodes.\\

The above bias is due to the well known bias in variance estimation, which can be eliminated with the bias correction (\ref{eq:PG0mod}), as outlined in the main text.
We now show that the bias for this modified Gini impurity is zero for OOB data.
As before, $\widehat{G}_{oob} = \widehat{PG}_{oob}^{(0)}$:
\begin{align*}
E[\Delta \widehat{PG}_{oob}(m)] &= E[\widehat{G}_{oob}(m)] - \frac{N_{m_l}}{N_{m}} E[\widehat{G}_{oob}(m_l)] - \frac{N_{m_r}}{N_{m}} E[\widehat{G}_{oob}(m_r)]  \\
& = \sigma^2_{m,oob} \cdot  \left[  1 - \frac{N_{m_l} + N_{m_r}}{N_{m}}  \right]  =  0
\end{align*}

\end{document}